\documentclass[11pt,a4paper]{article}
\usepackage[hyperref]{acl2020}
\usepackage{times}
\usepackage{latexsym}
\usepackage{booktabs}
\usepackage{multirow}
\usepackage{graphicx}

\usepackage{microtype}

\aclfinalcopy %

\usepackage{siunitx}

\newcommand{\x}{$X$}
\newcommand{\xs}{$X^{*}$}
\newcommand{\xss}{$X^{**}$}
\newcommand{\y}{$Y$}
\newcommand{\ys}{$Y^{*}$}
\newcommand{\yss}{$Y^{**}$}

\newcommand{\yswmt}{$Y^{*}_{WMT}$}

\newcommand{\bitext}{OP} %
\newcommand{\btnews}{BT}

\title{On The Evaluation of Machine Translation Systems \\ Trained With Back-Translation}

\author{Sergey Edunov \quad  Myle Ott \quad  Marc'Aurelio Ranzato \quad  Michael Auli \\
  Facebook AI Research %
}

\date{}

\begin{document}
\maketitle
\begin{abstract}
Back-translation is a widely used data augmentation technique which leverages target monolingual data. 
However, its effectiveness has been challenged since automatic metrics such as BLEU only show significant improvements for test examples where the source itself is a translation, or \emph{translationese}.
This is believed to be due to translationese inputs better matching the back-translated training data.
In this work, we show that this conjecture is not empirically supported and that back-translation improves translation quality of both naturally occurring text as well as translationese according to professional human translators. 
We provide empirical evidence to support the view that back-translation is preferred by humans because it produces \emph{more fluent} outputs. 
BLEU cannot capture human preferences because references are translationese when source sentences are natural text. 
We recommend complementing BLEU with a language model score to measure fluency.
\end{abstract}

\section{Introduction}

Back-translation (BT; \citealt{bojar:bt_pbmt:2011}; \citealt{sennrich2016bt}; \citealt{poncelas:investigatingBT:2018}) is a data augmentation
method that is a key ingredient for improving translation quality of neural machine translation systems (NMT; \citealt{sutskever2014seq2seq,bahdanau2015neural,gehring2017convs2s,vaswani2017transformer}). NMT systems using large-scale BT have been ranked top at recent WMT evaluation campaigns~\citep{bojar2018wmt,edunov2018bt,ng2019fairwmt}. 
The idea is to train a target-to-source model to generate 
additional synthetic parallel data from monolingual target data. 
The resulting sentence pairs have synthetic sources and natural targets which are then added to the original bitext in order to train the desired source-to-target model. BT improves generalization and can be used to adapt models to the test domain by adding appropriate monolingual data. 

Parallel corpora are usually comprised of two types of sentence-pairs: sentences which originate in the source language and have been translated by humans into the target language, or sentences which originate from the target language and have been translated into the source language.
We refer to the former as the \emph{direct} portion and the latter as the \emph{reverse} portion. 
The setup we are ultimately interested in is models that translate direct sentences.

Translations produced by human translators, or \emph{translationese} tend to be simpler and more standardized compared to naturally occurring text~\citep{baker1993translationese,zhang2019translationese,toury2012translationstudies}.
Several recent studies found that such reverse test sentences are easier to translate than direct sentences~\citep{toral2018wmt,graham2019translationese}, and
human judges consistently assign higher ratings to translations of target original sentences than to source original sentences. 
These studies therefore recommend to restrict test sets to source original sentences, a methodology which has been adopted by the 
2019 edition of the WMT news translation shared task.

Unfortunately, automatic evaluation with BLEU~\citep{papineni2002bleu} only weakly correlates with human judgements~\citep{graham2019translationese}.
Furthermore, recent WMT submissions relying heavily on back-translation mostly improved BLEU on the reverse direction with little gains 
on the direct portion~(\citealt{toral2018wmt}; Barry Haddow's personal communication and see also Appendix~\ref{app:wmt18}, Table~\ref{tab:wmt18sys}; \citealt{freitag2019textrepair}).

This finding is concerning for two reasons.
First, back-translation may not be effective after all since gains are limited to the reverse portion. 
Improvements on reverse sentences may only be due to a better match with the back-translated training sentences in this case.
Second, it may further reduce our confidence in automatic evaluation, if human judges disagree with BLEU for systems trained 
with back-translation. 
Indeed, human evaluations of top performing systems at WMT'18~\citep{bojar2018wmt} and WMT'19~\citep{bojar2019wmt} did not agree with BLEU to the extent that correlation is even negative for the top entries~\citep{ma2019metrics}.

In this paper, we shed light on the following questions. First, do BT systems only work better in the reverse direction? Second, does BLEU reflect human assessment for BT models? And if that is not the case, why not and how can we alleviate the weaknesses of BLEU? 

Our contribution is an extensive empirical evaluation of top-performing
NMT systems to validate or disproof some of the above conjectures. First, we show that translationese sources are indeed easier to translate, but this is 
true for both NMT systems trained with and without back-translated data. Second, we confirm that human assessment of BT systems poorly correlates with BLEU.
Third, BLEU cannot capture the higher quality of back-translation systems because the outputs of both back-translation and non back-translation models are equally close to the translationese references. 
Fourth, we show that BT system outputs are significanlty more
fluent than the output of a system only trained on parallel data, and this may explain the human preference towards BT generations. Finally, we recommend to improve automatic evaluation by complementing BLEU with a language model score which can better assess fluency in the target language while avoiding the artifacts of translationese references.

\section{Related Work}

Back-translation has been originally introduced for phrase-based machine translation~\citep{bojar:bt_pbmt:2011}. 
For back-translation with neural machine translation, there is a large body of literature building upon the seminal work of~\citet{sennrich2016bt},
from large-scale extensions with sampling~\citep{edunov2018bt,ott2018scaling} or tagging~\citep{caswell2019tagged} to its use for unsupervised machine translation~\citep{unsupMTLample} as well as analysis~\citep{poncelas2018investigatingBT} and iterative versions~\citep{hoang2018iterative}.

More similar to our work, \citet{toral2018wmt} analyzed performance of trained state-of-the-art NMT systems 
in direct and reverse mode. They observe that translationese is simpler to translate and claimed that gains for such systems mostly come from improvements 
in the reverse direction.

Concurrent to our work, \citet{graham2019translationese} find that automatic evaluation with BLEU does not align with the hypothesis 
that reverse sentences are easier to translate instead. Unfortunately, their findings are not very conclusive because they do not control 
for the change of actual content, as sentences in one direction 
may be extracted from documents which are just harder to translate. In this work we correct for this effect by comparing translations of  
source original sentences with their double translations. 
 \citet{graham2019translationese} also observe that BLEU does not reliably correlate with human judgements. While they consider a large variety of systems trained in various ways, we instead focus on the comparison between the same NMT system trained with and without back-translated data.

Earlier work on statistical machine translation models argued in favor of using source original data only to train translation models~\citep{kurokowa2009detection}, language models for translation~\citep{lembersky2011lms}, and to tune translation models~\citep{stymne2017tuning}.
All these studies base most of their conclusions on automatic evaluation with BLEU, which is problematic since BLEU is not  
reliable and this procedure may overly optimize towards translationese references.

\citet{freitag2019textrepair} proposed a post-editing method to turn translationese system outputs into more natural text. As part of their evaluation, they
also observed that human assessments poorly correlate with BLEU. 
While we confirm some of these observations, our goal is an in-depth analysis of the evaluation of NMT systems trained with back-translated data. We provide empirical evidence corroborating the hypothesis 
that the discrepancy between BLEU and human assessment is due to the use of translationese references, and we provide a constructive suggestion on how to better automatically evaluate models trained with BT.

\section{Experimental Setup} \label{sec:setup}
In the next sections we first discuss the datasets and models used. 
Then, we report BLEU evaluations showing a big discrepancy between the gains obtained by a BT system in forward versus reverse direction compared to a baseline trained only on parallel data.
This is followed by a series of hypotheses about the reasons for this discrepancy, and empirical studies in support or to disprove these hypotheses.
We conclude with a recommendation for how to better evaluate NMT systems trained with BT.
 
\subsection{Training Datasets}
We consider four language directions: English-German (En-De), German-English (De-En), English-Russian (En-Ru) and Russian-English (Ru-En).

For En-De, we train a model on the WMT'18 news translation shared task data.
We used all available bitext excluding the ParaCrawl corpus. 
We removed sentences longer than 250 words as well as sentence-pairs with a source/target length ratio exceeding 1.5. 
This results in 5.18M sentence pairs.
For back-translation, we use the same setup as the WMT'18 winning entry for this language pair which entails sampled back-translation of 226M German newscrawl sentences~\citep{edunov2018bt}.\footnote{WMT'18 models are available at \url{https://github.com/pytorch/fairseq/tree/master/examples/backtranslation} and we used a single model.}

For De-En, En-Ru, Ru-En we use all parallel data provided by the WMT'19 news translation task, including Paracrawl.
We remove sentences longer than 250 words as well as sentence-pairs with a source/target length ratio exceeding 1.5 and sentences which are not in the correct language~\citep{lui2012langid}. 
This resulted in 27.7M sentence-pairs for En-De and 26M for En-Ru. 

For the back-translation models we use the top ranked Facebook-FAIR systems of the WMT'19 news shared translation task.\footnote{WMT'19 models are available at \url{https://github.com/pytorch/fairseq/tree/master/examples/wmt19}}
The parallel data and pre-processing of those systems is identical to our baselines which are trained only on parallel data~\citep{ng2019fairwmt}. 
As monolingual data, the WMT'19 newscrawl data was filtered by langid, resulting in 424M English and 76M Russian monolingual sentences.
For En-De and De-En models use a joined byte-pair encoding (BPE; \citealt{sennrich:bpe:2016}) with 32K split operations, and for En-Ru and Ru-En separate BPE dictionaries for the source and target with 24K split operations.

\subsection{Sequence to Sequence Models}

We train models using the big Transformer implementation of fairseq~\citep{vaswani2017transformer,ott2019fairseq}. 
All our models are trained on 128 Volta GPUs, following the setup described in~\citet{ott2018scaling}.
For En-De we used single Transformer Big models without checkpoint averaging. 
For De-En and En-Ru we increased model capacity by using larger FFN size (8192) and we also used an ensemble of models trained with three different seeds.

In the remainder of this paper, we will refer to baseline NMT models trained {\bf o}nly on {\bf p}arallel data as {\bf \bitext{}}, and to models trained on both
parallel data and {\bf b}ack-{\bf t}ranslated data as {\bf \btnews{}}.

\begin{figure}
\centering
\includegraphics[width=0.7\linewidth]{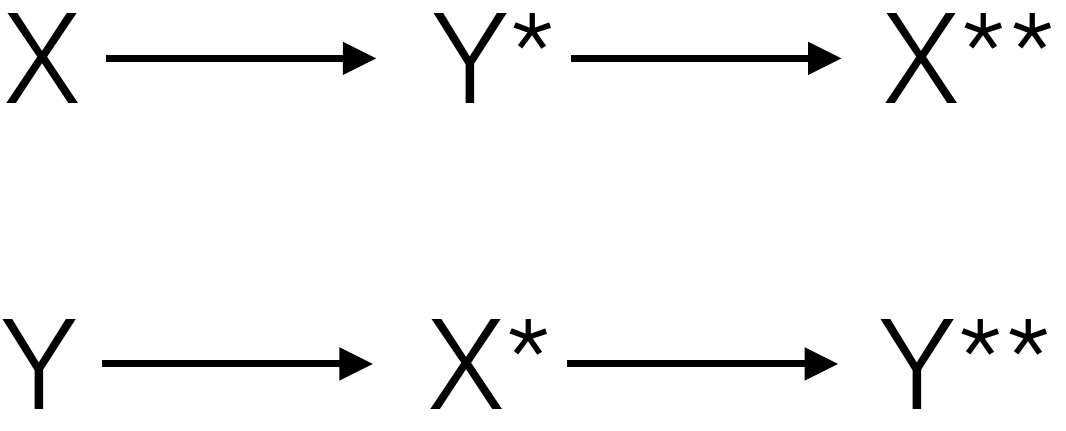}
\caption{Illustration of the translations used in this work. \x{} represent sentences originating in the source language. \y{} are sentences 
originating in the target language. A single $^*$ symbol represents a translation of an original sentence, while $^{**}$ represents a double translation, 
i.e. a translation of a translationese sentence. The original dataset consists of the union of (\x{}, \ys{}) pairs (direct mode) and (\xs{}, \y{}) 
(reverse mode).
According to BLEU, a system trained with BT improves only in reverse mode. As part of this study we have collected double translations, which are useful to 
assess whether translationese inputs are easier to translate 
(by comparing performance when the input is \xss{} versus \x{} and the reference is \ys{}) and easier to predict 
(by comparing performance when the reference is \yss{} versus \y{} and the input is \xs{}).
}
\label{fig:refs}
\end{figure}

\subsection{Test sets and Reference Collection}
\label{sec:testset}

In order to assess differences in model performance when inputting translationese vs. natural language (\textsection\ref{sec:simpleinp}), we collected additional references which 
will be made publicly and freely available soon.3
These are sentence-level (as opposed to document level) translations which matches the training setup of our models.
In Appendix~\ref{app:wmtref} we confirm that our findings also apply to the original WMT document-level references.

Figure~\ref{fig:refs} illustrates the composition of the test set for each language direction which is divided into two partitions: 
First, the \emph{direct} portion consists of sentences \x{} originally written in the source language which were translated into the target language as \ys{}. 
Additionally, we translated \ys{} back into the source language to yield \xss{}, a translationese version of \x{}.
Second, for the \emph{reverse} portion, we have naturally occurring sentences in the target language \y{} that were translated into the source as \xs{}. 
We also translated these into the target as \yss{} to obtain a translationese version of the original target.
For each language pair we use the following data:

\paragraph{English  $\leftrightarrow$ German.}
We used newstest2014 that we separated into English-original and German-original sets. We then sampled 500 English-original and 500 German-original sentences from each subset and asked professional human translators to translate them into German and English respectively. 
In addition, we ask professional human translators to provide \xss{} and \yss{} which are translations of \ys{} and \xs{}, respectively.

\paragraph{English $\leftrightarrow$ Russian.}
For this setup we sampled 500 English-original sentences from the En-Ru version of newstest2019 and asked professional human translators to translate them into Russian at 
the sentence-level. 
Similarly, we sampled 500 Russian-original sentences from the Ru-En version of newstest2019 and obtained English references.
We also collected double translations \xss{}, \yss{} of \ys{} and \xs{}, respectively.
3
The additional references are available at \url{https://github.com/facebookresearch/evaluation-of-nmt-bt}.

\subsection{Human and Automatic Evaluation}
\label{sec:humaneval}

Human evaluations and translations were conducted by certified professional translators who are native speakers of the target language and fluent in the source language.
We rate system outputs using both source and target based direct assessment. In the former case, raters evaluate correctness and completeness on a 
scale of 1-100 for each translation given a source sentence. 
This method is the most thorough assessment of translation quality. It also has the additional benefit to be independent
of the provided human references which may affect the evaluation. 
For target based direct assessment, raters evaluate closeness to the provided reference
on a scale of 1-100 for each translation. 
This is easier since it only requires people fluent in one language, and it is the evaluation performed by recent WMT campaigns~\citep{graham_baldwin_moffat_zobel_2017,bojar2018wmt}. 

To rate a translation, we collected three judgements per sentence. 
We repeated the evaluation for sentences where all three raters provided judgements that differed by more than 30 points.
Evaluation was blind and randomized: human raters did not know the identity of the systems and all outputs were shuffled to ensure that each rater provides a similar number of judgements for each system.

Following the WMT shared task evaluation~\citep{bojar2018wmt}, we normalize the scores of each rater by the mean and standard deviation of all ratings provided by the rater.
Next, we average the normalized ratings for each sentence and average all per-sentence scores to produce an aggregate per-system z-score.
As automatic metric, we report case-sensitive BLEU using SacreBLEU~\cite{matt2018clarity}.\footnote{SacreBLEU signature: BLEU+case.mixed+numrefs.1+\newline{}smooth.exp+tok.13a+version.1.3.1}
We also consider other metrics in Appendix~\ref{app:othermetrics}, but conclusions remain the same.

\begin{table}[t]
\centering
\begin{tabular}{p{5mm}p{5mm}p{5mm}rrrr}
\toprule
src & ref & sys & en-de & de-en & en-ru & ru-en \\
\midrule
\multirow{2}{*}{\x{}} & \multirow{2}{*}{\ys{}} 
& \bitext{} & 33.7 & 40.3 & 31.3 & 43.8 \\
& & \btnews{}& 32.3 & 38.6 & 31.9 & 41.2 \\
\midrule
\multirow{2}{*}{\xs{}} & \multirow{2}{*}{\y{}}
& \bitext{} & 31.3 & 43.0 & 40.5 & 31.8 \\
& & \btnews{} & 38.9 & 48.7 & 50.6 & 40.3 \\
\bottomrule
\end{tabular}
\caption{BLEU for four language directions measured on source original sentences (\x{} $\rightarrow$ \ys{}) as well as target original sentences (\xs{} $\rightarrow$ \y{}) for a model trained on parallel data only (\bitext{}) as well as a back-translation model (\btnews{}).
\btnews{} performs much better than \bitext{} on the reverse portion of the test set but BLEU shows no difference on the direct portion.
}
\label{tab:dirrev}
\end{table}

\begin{table}[t]
\centering
\begin{tabular}{p{5mm}p{5mm}p{5mm}rrrr}
\toprule
src & ref & sys & en-de & de-en & en-ru & ru-en \\
\midrule
\multirow{2}{*}{\x{}} & \multirow{2}{*}{\ys{}} 
& \bitext{} & 33.7 & 40.3 & 31.3 & 43.8 \\
& & \btnews{}& 32.3 & 38.6 & 31.9 & 41.2 \\
\midrule
\multirow{2}{*}{\xss{}} & \multirow{2}{*}{\ys{}} 
& \bitext{} & 39.7 & 46.9 & 42.8 & 49.9 \\
& & \btnews{} & 39.2 & 45.6 & 44.0 & 47.6 \\
\bottomrule
\end{tabular}
\caption{BLEU for source original sentences (\x{} $\rightarrow$ \ys{}) compared to the same sentence pairs with a translationese source (\xss{} $\rightarrow$ \ys{}). Translationese inputs are simpler to translate but \btnews{} and \bitext{} systems benefit equally from translationes inputs.
}
\label{tab:translationese}
\end{table}

\begin{table*}[t]
\centering
\begin{tabular}{lllrrrrrrrr}
\toprule
src & ref & sys & \multicolumn{2}{c}{en-de} & \multicolumn{2}{c}{de-en} & \multicolumn{2}{c}{en-ru} & \multicolumn{2}{c}{ru-en} \\
& & & BLEU & human & BLEU & human & BLEU & human & BLEU & human\\
\midrule
\multirow{2}{*}{\x{}} & \multirow{2}{*}{\ys{}} 
& \bitext{} & 33.7 & -0.18 & 40.3 & -0.07 & 31.3 & -0.66   & 43.8 & -0.37 \\
& & \btnews{}& 32.3 & \bf{-0.05} & 38.6 & \bf{0.03} & 31.9 & \bf{-0.35}  & 41.2 & \bf{-0.12} \\
\midrule
\multirow{2}{*}{\xs{}} & \multirow{2}{*}{\y{}}
& \bitext{} & 31.3 & -0.01 & 43.0 & 0.06 & 40.5 & 0.06 & 31.8 & -0.02 \\
& & \btnews{} & 38.9 & \bf{0.10} & 48.7 & \bf{0.13} & 50.6 & \bf{0.16} & 40.3 & \bf{0.07} \\
\midrule
\multirow{2}{*}{\xss{}} & \multirow{2}{*}{\ys{}} 
& \bitext{} & 39.7 & -0.05 & 46.9 & 0.07 & 42.8 & -0.17 & 49.9 & -0.05 \\
& & \btnews{} & 39.2 &  \bf{0.03} & 45.6 & \bf{0.16} & 44.0 & \bf{-0.01} & 47.6 & \bf{0.12} \\
\midrule
\multirow{2}{*}{\xs{}} & \multirow{2}{*}{\yss{}}
&   \bitext{} & 39.5 & -0.01 & 63.6 & 0.06 & 49.5 & 0.06  & 44.4 & -0.02 \\
& & \btnews{} & 41.8 & \bf{0.10} & 61.2 & \bf{0.13} & 50.4 & \bf{0.16} & 38.7 & \bf{0.07} \\
\bottomrule
\end{tabular}
\caption{BLEU and human preference judgements on four language directions with a bitext-only model as well as a back-translation model (\btnews{}).
BLEU shows no strong preference when the source is natural text (\x{}) but professional human translators prefer \btnews{} regardless of whether the source is \x{} or translationese (\xs{}).
Back-translation also does not overproportionally benefit from inputting translationese since both \bitext{} and \btnews{}  show similar improvements when switching from \x{} to \xss{} inputs. \btnews{} human scores are statistically significantly better at p=0.05 than the respective \bitext{} as per paired bootstrap resampling~\citep{koehn2004bootstrap}. 
}
\label{tab:human}
\end{table*}

\section{Results}

\subsection{Evaluating BT with Automatic Metrics}
We first reproduce the known discrepancy between \btnews{} and \bitext{} in the reverse direction 
(target original sentences; \xs{} $\rightarrow$ \y{}) and the forward direction (source original sentences; \x{} $\rightarrow$ \ys{}).

Table~\ref{tab:dirrev} shows that \btnews{} does not improve over \bitext{} on direct sentences (\x{} $\rightarrow$ \ys{}) in aggregate. However, on the reverse portion \btnews{} does improve, and it does so by very large margins of between 5.7-10.1 BLEU.
Appendix~\ref{app:othermetrics} shows that TER~\citep{snover2006ter}, BEER~\citep{stanojevic2014beer}, METEOR~\citep{banerjee2005meteor} and BERTScore~\citep{zhang2019bertscore} also do not distinguish very strongly between \bitext{} and \btnews{} for direct sentences.

A possible explanation for this result is that \btnews{} can better translate target-original test sentences because those sentences mimic the training data of 
\btnews{}. 
The \btnews{} training data (\textsection\ref{sec:setup}) consists largely of target original sentences-pairs with back-translated sources which could explain the discrepancy between performance of the \btnews{} system on the direct and reverse portions.

\subsection{Translationese Benefits Both \btnews{} \& \bitext{}} 
\label{sec:simpleinp}

Translationese is known to be a different dialect with lower complexity than naturally occurring text~\citep{toral2018wmt}. 
This is corroborated by the fact that this data is straightforward to identify by simple automatic classifiers~\citep{koppel2011translationese}.
One possible explanation for why back-translation could be more effective for target original sentences is that the input to the system is 
translated language. This may give the BT system two advantages: 
i) the input is simpler than naturally occurring text and 
ii) this setup may be easier for the back-translation system which was trained on additional target original data that was automatically translated.

To test this hypothesis we feed source original sentences and translationese into our systems and compare their performance. 
We created a test setup where we have \emph{both} a source original sentence (\x{}) and a translationese version of it (\xss{}) which share a reference (\y{}), see \textsection\ref{sec:testset}.
This enables us to precisely test the effect of translationese vs natural language. 

Table~\ref{tab:translationese} shows that BLEU is substantially higher when the input is translationese (\xss{}) compared to natural language (\x{}), 
however, both \btnews{} and \bitext{} obtain comparable improvements.
Therefore, the BLEU discrepancy between \btnews{} and \bitext{} in direct vs. reverse cannot be explained  by \btnews{} gaining an advantage over \bitext{} through translationese inputs.

\subsection{Human Evaluation Contradicts BLEU}

The aforementioned experiments were evaluated in terms of BLEU, an automatic metric. 
To get a more complete picture, we ask professional human translators to judge translations using source-based direct assessment (unless otherwise specified, this is our default type of human evaluation; see \textsection\ref{sec:humaneval}).

Table~\ref{tab:human} (first two sets of rows) shows that human judges prefer \btnews{} over \bitext{} regardless of whether sentences are source original (\x{} $\rightarrow$ \ys{}) or target original (\xs{} $\rightarrow$ \y{}). 
This is in stark contrast to the corresponding BLEU results. 

Similar observations have been made in the two most recent WMT evaluation campaigns: at WMT'18~\citep{bojar2018wmt}, the large-scale sampled \btnews{} system of Facebook-FAIR~\citep{edunov2018bt} ranked 6th in terms of BLEU while being ranked first in the human evaluation. 
The results of WMT'19 show a similar picture where a system relying on large scale back-translation ranked first in the human evaluation but only 8th in terms of BLEU~\citep{bojar2019wmt}.

\textit{We conclude that professional human translators prefer \btnews{} over \bitext{} - regardless of whether test sentences are source or target original.}

\subsection{Human Evaluation is Robust}

Our current observations could be explained by some idiosyncrasy in the human evaluation. 
To reject this hypothesis we performed both source-based and target-based assessment for all English-German systems of Table~\ref{tab:human} using professional translators (\textsection\ref{sec:humaneval}) and computed the correlation between the two types of assessments.
The correlation coefficient between source and target based assessment is 0.90 (95\% confidence interval 0.55 - 0.98), which
indicates that human evaluation is robust to the assessment type. 
This finding is consistent with other work comparing the two types of human evaluations~\citep{bojar2018wmt}.

\subsection{Why BLEU Fails in Direct Mode} 
\label{sec:bleu_broken}

Next, we investigate why BLEU does not agree with human judgements in direct mode.
BLEU measures n-gram overlap between a model output and a human reference translation.
In the case of direct sentences, the references are translationese.

We found earlier that BLEU does not distinguish between \btnews{} and \bitext{} even though professional human translators prefer \btnews{}.
Given references are translationese, one possible explanation is that both systems produce translations which equally resemble translationese 
and thus BLEU fails to distinguish between them.

To test this hypothesis and measure the closeness of system outputs with respect to translationese, we train two large transformer-based language 
models~\citep{baevski2018adp}.
The first is trained on outputs produced by the En-De \btnews{} system, the second one on the outputs produced by 
the En-De \bitext{} system. The outputs are the translation of English Newscrawl 2018 comprising 76M sentences.
We then evaluate the language models on source original sentences (\ys{}) of newstest2015-2018.

The first row of Table~\ref{tab:lms} shows that both language models achieve similar perplexity on \ys{} (37.2 VS 36.8), suggesting that the translations of \btnews{} and \bitext{} are equally close to translationese.
Interestingly, both system outputs are closer to translationese than natural text since PPL on \ys{} is significantly lower than the PPL on \y{} (second row of Table~\ref{tab:lms}). 
This is also supported by BLEU being higher when using \yss{} as a reference compared to \y{} for the same input \xs{} (second and last row of Table~\ref{tab:human}).

\emph{Our results support the hypothesis that the outputs of \btnews{} and \bitext{} are equally close to translationese. This in turn may explain why BLEU cannot distinguish between \bitext{} and \btnews{} in direct mode where the reference is translationese.}

\begin{table}
\centering
\begin{tabular}{lrr}
\toprule
data     & \bitext{} & \btnews{} \\
\midrule
\ys{} & 37.2 & 36.8 \\
\y{}     & 82.2 & 57.4 \\
\bottomrule
\end{tabular}
\caption{Perplexity on the source-original/translationese portion (\ys{)} and the target-original portion of newstest2014-2018 (\y{}). 
We translate the English newscrawl training data with either \bitext{} and \btnews{} and train two language models on the outputs. 
Both \btnews{} and \bitext{} are equally close to translationese (first row), but \btnews{} is closer than \bitext{} to naturally occurring text 
(second row).}
\label{tab:lms}
\end{table}

\subsection{BT Generates More Natural Text}
\label{sec:close2g}

Back-translation augments the training corpus with automatic translations from target original data.
Training models on large amounts of target original data may bias \btnews{} systems to produce outputs that are closer to naturally occurring text.
In contrast, \bitext{} systems have been trained on the original parallel data, a mix of direct and reverse data which contains a much smaller amount of target original sentences.
This may explain why BLEU evaluation with translationese references (direct portion) does not capture the human preference for \btnews{}. 

To understand this better, we conduct two experiments.
The first experiment is based on the language models we trained previously (\textsection\ref{sec:bleu_broken}) to assess how close our systems are to translationese and naturally occurring text. 
The second experiment is based on a human study where native speakers assess the fluency of each system output.

For the first experiment we reuse the two language models from \textsection\ref{sec:bleu_broken} to measure how close the system outputs are to natural text (\y).
The second row of Table~\ref{tab:lms} shows that the \btnews{} language model assigns much higher probability to naturally occurring text, \y, compared to the \bitext{} language model (82.2 VS 57.4 perplexity), suggesting that \emph{\btnews{} does indeed produce outputs that are much closer to natural text than \bitext{}}. 
We surmise that this difference, which is captured by a language model trained on system outputs and evaluated on \y{}, 
could be at least partially responsible for the marked human preference towards \btnews{} translations. 

In the second experiment, native speakers of English, German and Russian rate whether the output of \bitext{} is more fluent than the output of \btnews{} for 100 translations of the De-En, En-De and En-Ru systems. 
Human raters perform a pair-wise ranking and raters can only see two translations but not the source; the system identity is unknown to raters.

\begin{table}
\centering
\begin{tabular}{lrrr}
\toprule
& \btnews{} & \bitext{} & draw \\
\midrule
De-En & 28 & 16 & 63 \\
En-De & 50 & 33 & 18 \\
En-Ru & 37 & 21 & 42 \\
\bottomrule
\end{tabular}
\caption{Human preference in terms of fluency for system outputs of \btnews{} and \bitext{}. Judgements are based on a pair-wise comparison between the two systems without the source sentence and conducted by native speakers. 
All results are based on 100 judgements and the preference of BT over OP is statistically significant at p=0.05.}
\label{tab:human_fluency}
\end{table}

Table~\ref{tab:human_fluency} shows that \btnews{} is judged to be significantly more fluent by native speakers than \bitext{} in three languages.

\section{Improving \btnews{} Evaluation} \label{sec:recomm}

In the previous sections, we gathered mounting evidence that BLEU fails at capturing the improved fluency of \btnews{} in direct mode. 
Next, we propose to use a language model to assess fluency as an additional measure to complement BLEU.
Different to the setup above (\textsection\ref{sec:bleu_broken}, \ref{sec:close2g}), where we used a separate LM for each system, we propose 
to use {\em a single} LM for all systems in order to simplify the evaluation. 

The language model is trained on a large monolingual dataset {\em disjoint} from the monolingual dataset used for generating back-translated data for \btnews{} training. 
This restriction is critical, otherwise the language model is likely to assign higher probably to \btnews{} generations simply because training and evaluation sets overlap.
To train these language models we sample 315M, 284M and 120M commoncrawl sentences for each of the three target languages, namely English, German and Russian, respectively.

The language model is used to score the outputs of \btnews{} and \bitext{} on the direct portion of the test set. 
If two systems have similar BLEU scores, then a lower perplexity with the LM indicates higher fluency in the target natural language. 
This fluency assessment is complementary to BLEU which in turn is more sensitive to adequacy.

\begin{table}
\centering
\begin{tabular}{lrr}
\toprule
& \btnews{} PPL & \bitext{} PPL \\
\midrule
De-En & 74.8 & 78.7 \\
En-De & 48.6 & 52.6 \\
Ru-En & 57.6 & 68.6 \\
En-Ru & 61.7 & 72.4 \\
\bottomrule
\end{tabular}
\caption{Automatic fluency analysis with language models trained on the Common Crawl corpus in the respetive target language.
\btnews{} receives lower perplexity (PPL) throughout, despite attaining the same BLEU score of \bitext{}, see Table~\ref{tab:dirrev}.}
\label{tab:ccrawl_lm}
\end{table}

Table~\ref{tab:ccrawl_lm} shows that the language model assigns lower perplexity to \btnews{} in all four setups. 
This shows that a language model can help to assess the fluency of system output when a human evaluation is not possible.

In future work, we intend to further investigate how to best combine BLEU and language model scoring in order to maximize correlation with human
judgements, particularly when evaluating \btnews{} in direct mode. Meantime, practitioners can use this additional metric in their evaluation to break 
ties in BLEU scoring.

\section{Conclusions}
According to our findings, back-translation improves translation accuracy, for both source and target original sentences.
However, automatic metrics like BLEU fail to capture human preference for source original sentences (direct mode).

We find that \btnews{} produces outputs that are closer to natural text than the output of \bitext, which may explain human preference for \btnews.
We recommend distinguishing between direct and reverse translations for automatic evaluation, and to make final judgements based on human evaluation.
If human evaluation is not feasible, complementing standard metrics like BLEU with a language model (\textsection\ref{sec:recomm}) 
may help assessing the overall translation quality.

In the future, we plan to investigate more thoroughly the use of language models for evaluating fluency, the effect of domain mismatch in the 
choice of monolingual data, and ways to generalize this study to other applications beyond MT.

\section*{Acknowledgements}

We thank Barry Haddow for initially pointing out the BLEU discrepancy between the forward and reverse portions of the WMT 2018 test set.

\bibliography{master}
\bibliographystyle{acl_natbib}

\clearpage
\onecolumn
\appendix

\section{Forward/reverse BLEU for WMT'18 English-German systems} \label{app:wmt18}
\begin{table*}[!h]
\centering
\begin{tabular}{lrrr}
\toprule
system & fwd & rev & delta \\
\hline
online-Y & 47.1 & 30.3 & -16.8 \\
MMT-production-system & 51.8 & 36.7 & -15.1 \\
online-B.0 & 52.9 & 39.1 & -13.8 \\
NTT & 50.7 & 39.7 & -11.0 \\
Microsoft-Marian & 52.5 & 41.6 & -10.9 \\
KIT & 50.3 & 39.5 & -10.8 \\
LMU-nmt & 43.5 & 33.4 & -10.1 \\
uedin & 47.8 & 37.8 & -10.0 \\
online-A & 37.8 & 28.6 & -9.2 \\
JHU & 46.0 & 38.2 & -7.8 \\
online-F & 23.5 & 16.4 & -7.1 \\
UCAM & 48.9 & 42.1 & -6.8 \\
RWTH-UNSUPER & 16.7 & 12.0 & -4.7 \\
online-G & 25.9 & 22.5 & -3.4 \\
LMU-unsup & 15.2 & 14.3 & -0.9 \\
\textbf{Facebook-FAIR} & \textbf{45.8} & \textbf{46.1} & \textbf{0.4} \\
\bottomrule
\end{tabular}
\caption{Forward/reverse BLEU for WMT'18 English-German systems.}
\label{tab:wmt18sys}
\end{table*}

Table~\ref{tab:wmt18sys} shows that a large-scale back-translation system, Facebook-FAIR, mostly improves BLEU on the reverse portion whereas it is outperformed by many other entrants in the forward portion.

\section{Results with WMT references} \label{app:wmtref}
\begin{table*}[!h]
\centering
\begin{tabular}{lllrrrrrrrr}
\toprule
src & ref & sys & \multicolumn{2}{c}{en-de} & \multicolumn{2}{c}{de-en} & \multicolumn{2}{c}{en-ru} & \multicolumn{2}{c}{ru-en}\\
& & & BLEU & human & BLEU & human & BLEU & human & BLEU & human\\
\midrule 
\multirow{2}{*}{\x{}} & \multirow{2}{*}{\ys{}} 
& \bitext{} & 33.7 & -0.18 & 40.3 & -0.07 & 31.3 & -0.66 & 43.8 & -0.37 \\
& & \btnews{}& 32.3 & -0.05 & 38.6 & 0.03 & 31.9 & -0.35 & 41.2 & -0.12 \\
\midrule
\multirow{2}{*}{\x{}} & \multirow{2}{*}{\yswmt{}}
& \bitext{} & 28.7 & -0.18 & 35.4 & -0.07 & 31.8 & -0.66  & 39.7 & -0.37 \\
& & \btnews{} & 29.9 & -0.05 & 34.2 & 0.03 & 31.9 & -0.35 & 38.5 & -0.12 \\
\bottomrule
\end{tabular}
\caption{BLEU results with respect to the original WMT references (document-level) and the sentence-level references used throughout this study. 
Sentence-level references result in higher BLEU but \bitext{} and \btnews{} still achieve very similar BLEU.
}
\label{tab:wmtref}
\end{table*}

Table~\ref{tab:wmtref} shows that BLEU does not strongly distinguish between \btnews{} and \bitext{}, regardless of whether the reference was obtained at the document-level (\yswmt{}) or at the sentence-level (\ys{}).

\clearpage
\section{Other metrics than BLEU}
\label{app:othermetrics}

\begin{table*}[!h]
\centering
\begin{tabular}{lllrrrrrrrrrrrr}
\toprule
src & ref & sys & \multicolumn{6}{c}{en-de} \\
& & & human & BLEU & TER & BEER & METEOR & BERTScore \\
\midrule
\multirow{2}{*}{\x{}} & \multirow{2}{*}{\ys{}}
&   \bitext{}    & -0.18 & 33.7 & 0.466 & 0.635 & 0.531 & 0.849 \\
& & \btnews{}    & -0.05 & 32.3 & 0.473 & 0.619 & 0.512 & 0.843\\
\midrule
\multirow{2}{*}{\xs{}} & \multirow{2}{*}{\y{}}
& \bitext{}      & -0.01  & 31.3 & 0.504 & 0.609 & 0.530 & 0.841 \\
& & \btnews{}    & 0.10 & 38.9 & 0.431 & 0.652 & 0.580 & 0.866 \\
\midrule
\multirow{2}{*}{\xss{}} & \multirow{2}{*}{\ys{}} 
& \bitext{}      & -0.05 & 39.7 & 0.403 & 0.677 & 0.590 & 0.878 \\
& & \btnews{}    & 0.03 & 39.2 & 0.409 & 0.669 & 0.578 & 0.876 \\
\midrule
\multirow{2}{*}{\xs{}} & \multirow{2}{*}{\yss{}}
&   \bitext{}    & -0.01  & 39.5 & 0.410 & 0.670 & 0.599 & 0.876 \\
& & \btnews{}    & 0.10 & 41.8 & 0.383 & 0.683 & 0.610 & 0.884 \\
\bottomrule
\end{tabular}
\caption{BLEU and other metrics as well as human preference judgements for English-German translations.
}
\label{tab:metrics}
\end{table*}

Table~\ref{tab:metrics} shows results for automatic metrics other than BLEU~\citep{papineni2002bleu}. 
The metrics TER~\citep{snover2006ter}, BEER~\citep{stanojevic2014beer}, METEOR~\citep{banerjee2005meteor} and BERTScore~\citep{zhang2019bertscore} show similar trends as BLEU, i.e., they do not indicate human preference of BT over bitext for the direct portion of the test set (\x{} $\rightarrow$ \ys{}).

\end{document}